\documentclass[conference]{IEEEtran}
\IEEEoverridecommandlockouts
\usepackage{amsmath,amssymb,amsfonts}
\usepackage{algorithmic}
\usepackage{graphicx}
\usepackage{textcomp}
\usepackage{xcolor}
\usepackage{hyperref}

\usepackage{listings}

\usepackage{xparse}
\usepackage{CJKutf8} 

\NewDocumentCommand{\codeword}{v}{%
\texttt{\textcolor{black}{#1}}%
}

\usepackage[numbers,square]{natbib}
\def\BibTeX{{\rm B\kern-.05em{\sc i\kern-.025em b}\kern-.08em
    T\kern-.1667em\lower.7ex\hbox{E}\kern-.125emX}}
\begin{document}

\title{Black-Box Analysis: GPTs Across Time in Legal Textual Entailment Task}

\author{\IEEEauthorblockN{
    Ha-Thanh Nguyen\textsuperscript{1,*}\thanks{\textsuperscript{*} Corresponding: nguyenhathanh@nii.ac.jp},
    Randy Goebel\textsuperscript{2},
    Francesca Toni\textsuperscript{3},\\
    Kostas Stathis\textsuperscript{4},
    Ken Satoh\textsuperscript{1}
}
\IEEEauthorblockA{
\textit{\textsuperscript{1}National Institute of Informatics (NII), 2-1-2 Hitotsubashi, Chiyoda City, Tokyo, Japan}\\
\textit{\textsuperscript{2}University of Alberta, Alberta Machine Intelligence Institute, 116 St \& 85 Ave, Edmonton, AB T6G 2R3, Canada}\\
\textit{\textsuperscript{3}Imperial College London, South Kensington Campus, London SW7 2AZ, UK}\\
\textit{\textsuperscript{4}Royal Holloway, University of London, Egham Hill, Egham TW20 0EX, UK}
}}

\maketitle

\begin{abstract}
The evolution of Generative Pre-trained Transformer (GPT) models has led to significant advancements in various natural language processing applications, particularly in legal textual entailment. We present an analysis of GPT-3.5 (ChatGPT) and GPT-4 performances on COLIEE Task 4 dataset, a prominent benchmark in this domain. The study encompasses data from Heisei 18 (2006) to Reiwa 3 (2021), exploring the models' abilities to discern entailment relationships within Japanese statute law across different periods. Our preliminary experimental results unveil intriguing insights into the models' strengths and weaknesses in handling legal textual entailment tasks, as well as the patterns observed in model performance. In the context of proprietary models with undisclosed architectures and weights, black-box analysis becomes crucial for evaluating their capabilities. We discuss the influence of training data distribution and the implications on the models' generalizability.  This analysis serves as a foundation for future research, aiming to optimize GPT-based models and enable their successful adoption in legal information extraction and entailment applications.

\end{abstract}

\section{Introduction}
\begin{figure}[ht]
\centering
\includegraphics[width=.4\textwidth]{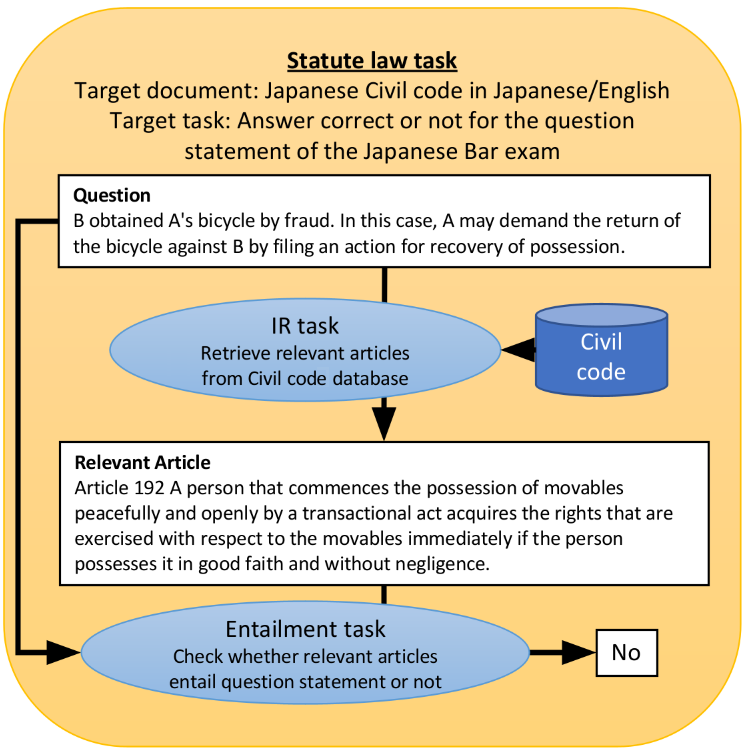}
\caption{An example of statute law retrieval (Task 3) and legal entailment (Task 4) in COLIEE Competition~\cite{Rabelo_2022}}
\label{fig:coliee}
\end{figure}

The Competition on Legal Information Extraction and Entailment (COLIEE) \cite{Rabelo_2020,Rabelo_2022} is a prominent platform for advancing research in automated legal reasoning, which holds significant importance as it enables a better understanding of complex legal documents and promotes the development of improved natural language processing (NLP) approaches. Task 4 in COLIEE specifically aims to extract critical information from Japanese legal text, such as statute law, and identify entailment relationships within the context of this domain. The objective of this task is to ascertain whether a given set of relevant articles (S1, S2, ..., Sn) entails a particular question (Q) or its negation (not Q), with the answers being binary, either ``YES'' (Q) or ``NO'' (not Q) (see Figure \ref{fig:coliee}).

By tackling this task, researchers can fine-tune their models to automatically extract pertinent information from legal documents and accurately identify entailment relationships, paving the way for more effective and efficient legal decision-making processes. This is particularly significant in a legal system like Japan, where qualifying as an attorney requires the completion of the Japanese Bar Examination, a stringent process entailing a law school curriculum, passing the bar examination, and successfully concluding an apprenticeship at the Legal Training and Research Institute.

Over the course of Generative Pre-trained Transformer (GPT) model development, substantial progress has been made in language understanding. GPT-2 \cite{radford2019language} showcased remarkable zero-shot learning capabilities on various language modeling tasks. Subsequently, GPT-3 \cite{brown2020language} substantially expanded the model size, leading to enhanced performance across a wide range of NLP tasks, such as translation and question-answering. The GPT-3.5 (also known as ChatGPT) \cite{ouyang2022training} model employed reinforcement learning from human feedback (RLHF), a fine-tuning strategy tailored to align the model more closely with user intent. Finally, GPT-4 \cite{openai2023gpt4} represents a multimodal model proficient in processing both image and text inputs, displaying human-like performance on several professional and academic benchmarks.

\begin{figure}[ht]
\centering
\includegraphics[width=.45\textwidth]{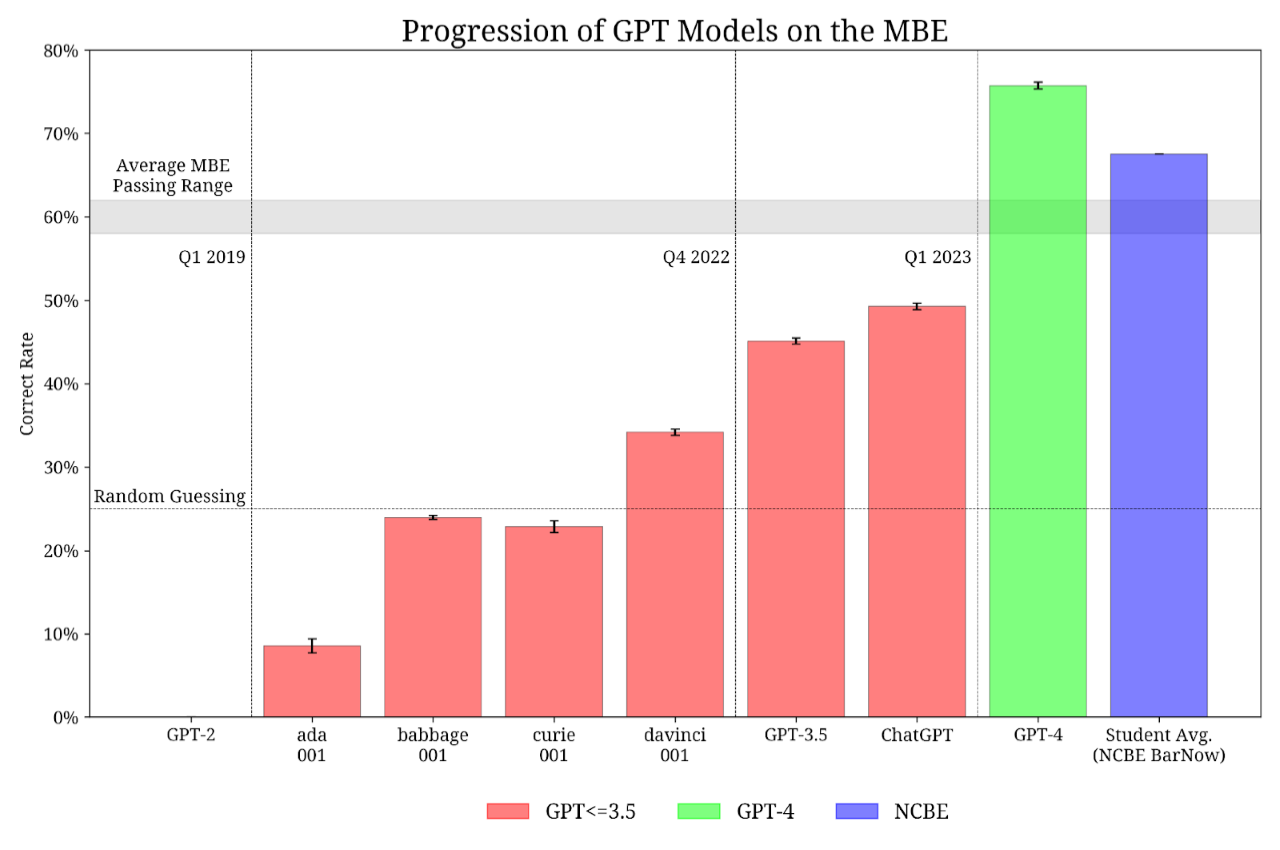}
\caption{GPT-4 significantly exceeds its predecessor models, as well as overtaking the average performance of law students on the MBE test. \cite{katz2023gpt}}
\label{fig:gpt4_mbe}
\end{figure}

GPT models have made remarkable strides in legal tasks, showcasing their potential to aid legal services across society. For example, GPT-4 has demonstrated exceptional performance in passing the Uniform Bar Examination (UBE), achieving a score that surpasses the passing threshold for all aspects, even exceeding the performance of human examinees in the Multistate Bar Examination (MBE)
(see Figure \ref{fig:gpt4_mbe}) 
\cite{katz2023gpt}. Furthermore, GPT-4 consistently outperforms ChatGPT in the same study.
Bilgin et al. \cite{bilgin2023amhr} also report impressive results with few-shot prompting approaches using GPT-4 on COLIEE 2023 Task 4 dataset.

In addition to GPT-4, the previous version GPT-3 has also been evaluated in the legal domain, specifically in the context of statutory reasoning tasks, a vital legal skill involving analyzing facts and rules written in natural language by legislatures \cite{blair2023can}. The study employed various approaches with the GPT-3 model, \codeword{text-davinci-003}, on a renowned statutory-reasoning dataset called SARA. While GPT-3 achieved impressive results, it exhibited limitations, making several clear errors due to its imperfect prior knowledge of the actual U.S. statutes and poor performance on synthetic statutes it hadn't encountered during training. Recent studies have also revealed limitations of language models in general, and GPTs in particular, in legal reasoning tasks \cite{nguyen2023well, nguyen2023beyond, nguyen2023negation}.

Given the advancements of GPT models in legal tasks and the identified limitations in previous versions, our study aims to compare the performance of GPT-3.5 (ChatGPT) and GPT-4 on the COLIEE Task 4 dataset, which focuses on legal textual entailment in the context of Japanese statute law. We seek to understand how these models address legal textual entailment tasks while exploring potential improvements in this important domain.

\section{Related Work}

The legal domain has always been a challenging and engaging area for researchers in the field of NLP. The complexity of legal language and the importance of contextual understanding in legal reasoning have given rise to numerous research studies to better equip NLP models in handling legal tasks.

In the domain of legal textual entailment, COLIEE Task 4 \cite{Rabelo_2020,Rabelo_2022} has been a vital benchmark dataset. Several approaches have been employed in order to tackle this task. Traditional NLP techniques, such as keyword-based matching \cite{kano2015keyword}, and rule-based systems \cite{rosa2021yes} have been used for information extraction and entailment identification in legal documents. These techniques, however, can suffer from rigid methodologies and lack the flexibility to handle subtle changes in language patterns and context.

The emergence of word embeddings like Word2Vec \cite{mikolov2013efficient} and transformers, such as BERT (Bidirectional Encoder Representations from Transformers) \cite{devlin2018bert}, RoBERTa \cite{liu2019roberta}, and ALBERT \cite{lan2019albert} models, revolutionized the field of NLP, leading to improvements in various legal tasks. Studies focusing on tasks like legal lawfulness classification \cite{nguyen2019deep}, information extraction \cite{zin2023improving}, question-answering \cite{kien2020answering} and multi-task learning \cite{vuong2023nowj} reported substantial progress due to these techniques.

Moreover, few-shot learning and few-relational learning strategies have been introduced to tackle legal entailment tasks \cite{bilgin2023amhr,rosa2022billions}. This enables models to learn from a limited amount of labeled data to generalize to unseen cases, which is a valuable approach given the scarcity of high-quality annotated legal documents.

Given the advancements offered by transformer models in handling many NLP tasks, they have been employed for legal textual entailment as well. For instance, BERT and its variants have shown promising results on the COLIEE dataset \cite{nguyen2020jnlp,nguyen2022transformer}, demonstrating their effectiveness in understanding complex legal language.

In recent years, the GPT family of models has emerged as a strong contender in various NLP tasks. These models have been deployed to tackle statutory reasoning tasks \cite{blair2023can,bilgin2023amhr}, revealing their usefulness in legal domains. With the incremental improvements in GPT models, GPT-4 has shown remarkable performance on the Uniform Bar Examination (UBE) \cite{katz2023gpt}, further highlighting the potential applications of GPT models in legal reasoning tasks. This research builds upon these breakthroughs, aiming to assess the performance of GPT-3.5 and GPT-4 in the legal textual entailment task within the context of Japanese statute law. We focus on the historical perspective by analyzing the data and models' performance from Heisei 18 (2006) to Reiwa 3 (2021), as this approach is crucial for understanding and anticipating the possible challenges that arise in the complex GPT models.

\section{Experimental Design}

In this section, we present the experimental design utilized in our study, which includes data preparation, interaction with the GPT models via API, and the evaluation of obtained results. The primary goal of our experiments is to assess the performance of GPT-3.5 (ChatGPT) and GPT-4 in Japanese legal textual entailment tasks using the COLIEE Task 4 dataset.

\subsection{Data Analysis}

\begin{figure}[ht]
\centering
\includegraphics[width=.45\textwidth]{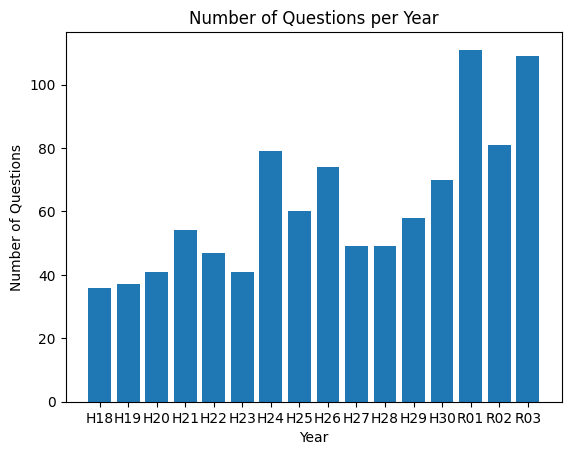}
\caption{The number of questions in the COLIEE Task 4 dataset in each year from H18 (2006) to R03 (2021).}
\label{fig:num_question}
\end{figure}

We examine the dataset used in our experiments, focusing on key statistics and trends that may affect the models' performance. The dataset contains questions along with their associated relevant articles from Japanese statute law spanning from Heisei 18 (2006) to Reiwa 3 (2021). Figure \ref{fig:num_question} depicts the yearly distribution of the dataset, with an increasing number of questions in recent years.

\begin{figure}[ht]
\centering
\includegraphics[width=.45\textwidth]{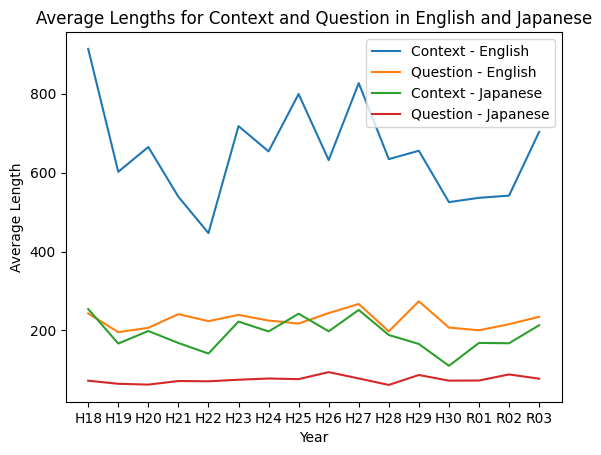}
\caption{Average length of context (related articles) and question in English and Japanese for each year in the COLIEE Task 4 dataset.}
\label{fig:avg_length}
\end{figure}

To better understand the dataset's characteristics, we analyze the average lengths of both context (related articles) and question in English and Japanese versions across different years. Figure \ref{fig:avg_length} showcases these trends in a line chart, revealing variations in the lengths of context and question over time.

The data indicate ranges of average lengths for context in English and Japanese as follows:
\begin{itemize}
\item English context length varies from 446 characters (H22) to 913 characters (H18).
\item Japanese context length ranges from 110 characters (H30) to 254 characters (H18).
\end{itemize}

English question average lengths fluctuate between 195 characters (H19) and 273 characters (H29), while Japanese question lengths range from 62 characters (H20, H28) to 94 characters (H26). Analyzing these variations in average lengths of both context and question in different years is important since it provides insights into the dataset's complexity and potential text processing challenges.

While these differences in context and question lengths show variability in the dataset, it's worth noting that advanced language models like GPT-3.5 and GPT-4 can handle thousands of tokens as input. As such, they should be capable of coping with the variations in the dataset's average lengths.

\subsection{GPT Model Interaction via API}

To conduct the experiments, we interacted with the GPT-3.5 (ChatGPT) and GPT-4 models by making API calls. For each instance in the dataset, we formatted the input prompt as follows in both English and Japanese:

\noindent \textbf{Prompt in English:}
\begin{quote}
\texttt{\{context\}\\
Question: \{question\}\\
Answer (Y or N), no explain.}
\end{quote}

\noindent \textbf{Prompt in Japanese:}
\begin{quote}
\texttt{\{context\}\\
\begin{CJK*}{UTF8}{min}質問\end{CJK*}: \{question\}\\
\begin{CJK*}{UTF8}{min}回答 (Y または N)、説明は不要。\end{CJK*}}
\end{quote}

Here, \texttt{context} represents the relevant articles, and \texttt{question} corresponds to a given question in the dataset. This prompt format ensures consistency in the input and allows the models to focus solely on providing a binary answer—either ``Y'' (YES) or ``N'' (NO). 

\subsection{Evaluation of Model Performances}

After receiving the predictions from the GPT models, we proceeded to evaluate their performances by comparing the predicted answers against the ground truth labels in the dataset. To align with the evaluation metric used by the COLIEE organizers, only accuracy was used as the primary metric to assess each model's ability to discern entailment relationships within Japanese statute law. This ensured a fair comparison and understanding of the models' strengths and weaknesses.

By following this experimental design, we effectively compared the performance of GPT-3.5 (ChatGPT) and GPT-4 on the legal textual entailment task using the COLIEE Task 4 dataset, paving the way for further analysis and discussion.
In the next section, we will dive into the experiments and present our findings from the analysis of the models' performances.

\section{Experimental Results}
In this section, we present a summary of the experimental results for GPT-3.5 (ChatGPT) and GPT-4 addressing the legal textual entailment task using the COLIEE Task 4 dataset. The results are organized into a table that provides an overview of the models' performance across different years, for both English and Japanese datasets. A detailed breakdown of various aspects of the results can be found in the corresponding figures, which visualize the accuracy scores and showcase a comprehensive comparison between the models.

\begin{figure}
\centering
\includegraphics[width=.4\textwidth]{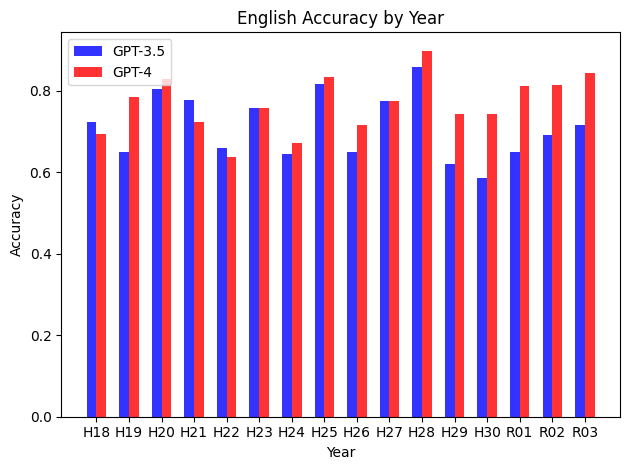}
\caption{Accuracy comparison between GPT-3.5 and GPT-4 with English data}
\label{fig:acc_english}
\end{figure}

\begin{figure}
\centering
\includegraphics[width=.4\textwidth]{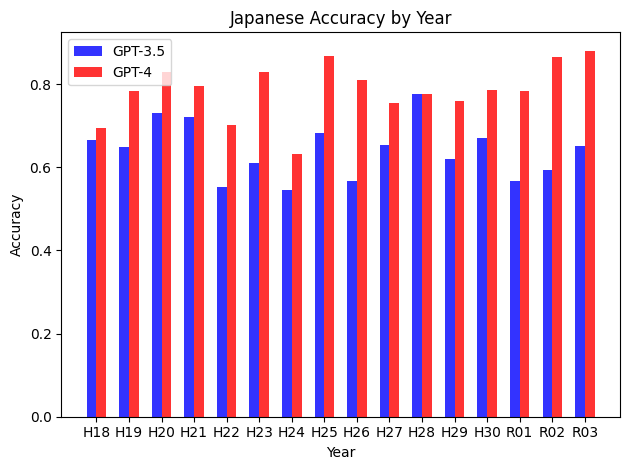}
\caption{Accuracy comparison between GPT-3.5 and GPT-4 with Japanese data}
\label{fig:acc_japanese}
\end{figure}

\begin{figure}
\centering
\includegraphics[width=.4\textwidth]{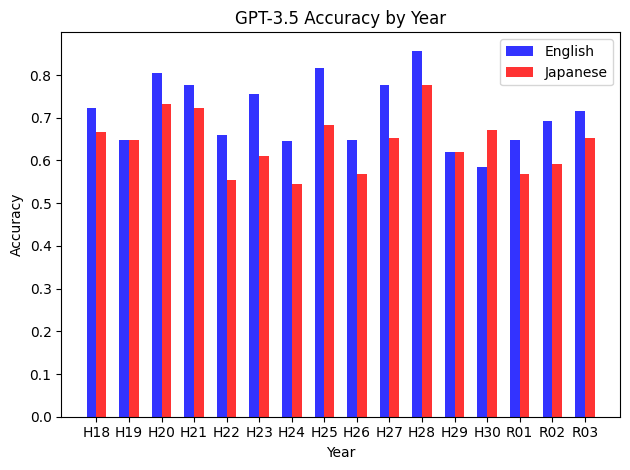}
\caption{Accuracy comparison of GPT-3.5 with English and Japanese data}
\label{fig:acc_gpt35_eng_jp}
\end{figure}

\begin{figure}
\centering
\includegraphics[width=.4\textwidth]{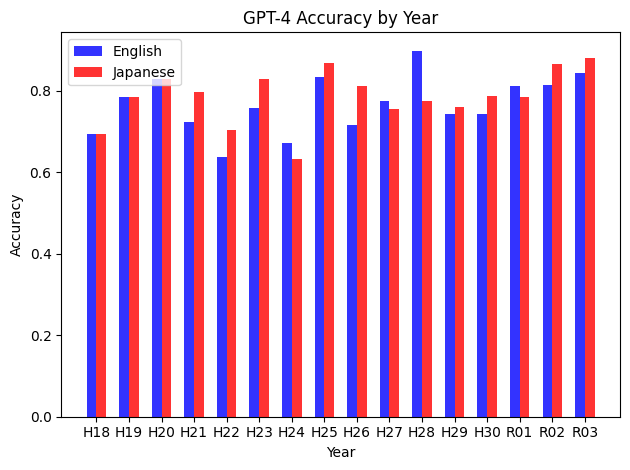}
\caption{Accuracy comparison of GPT-4 with English and Japanese data}
\label{fig:acc_gpt4_eng_jp}
\end{figure}

Table \ref{table:summary_results} presents a summary of the experimental results for GPT-3.5 and GPT-4 in both English and Japanese across the years. The table highlights each model's accuracy score for the English and Japanese data in the corresponding years. Figure \ref{fig:acc_english} illustrates the accuracy comparison between GPT-3.5 and GPT-4 with English data, while Figure \ref{fig:acc_japanese} showcases the same comparison with Japanese data. Figures \ref{fig:acc_gpt35_eng_jp} and \ref{fig:acc_gpt4_eng_jp} provide an in-depth look at the performance of GPT-3.5 and GPT-4, respectively, with both English and Japanese datasets.
\begin{table*}
\centering
\caption{Summary of GPT-3.5 and GPT-4 Results in English and Japanese}
\label{table:summary_results}
\begin{tabular}{|c|c|c|c|c|}
\hline
\textbf{Year} & \textbf{GPT-3.5 (English)} & \textbf{GPT-4 (English)} & \textbf{GPT-3.5 (Japanese)} & \textbf{GPT-4 (Japanese)} \\ \hline
H18         & 0.7222      & 0.6944      & 0.6667      & 0.6944      \\
H19         & 0.6486      & 0.7838      & 0.6486      & 0.7838      \\
H20         & 0.8049      & 0.8293      & 0.7317      & 0.8293      \\
H21         & 0.7778      & 0.7222      & 0.7222      & 0.7963      \\
H22         & 0.6596      & 0.6383      & 0.5532      & 0.7021      \\
H23         & 0.7561      & 0.7561      & 0.6098      & 0.8293      \\
H24         & 0.6456      & 0.6709      & 0.5443      & 0.6329      \\
H25         & 0.8167      & 0.8333      & 0.6833      & 0.8667      \\
H26         & 0.6486      & 0.7162      & 0.5676      & 0.8108      \\
H27         & 0.7755      & 0.7755      & 0.6531      & 0.7551      \\
H28         & 0.8571      & 0.8980      & 0.7755      & 0.7755      \\
H29         & 0.6207      & 0.7414      & 0.6207      & 0.7586      \\
H30         & 0.5857      & 0.7429      & 0.6714      & 0.7857      \\
R01          & 0.6486      & 0.8108      & 0.5676      & 0.7838      \\
R02          & 0.6914      & 0.8148      & 0.5926      & 0.8642      \\
R03          & 0.7156      & 0.8440      & 0.6514      & 0.8807      \\ \hline
\end{tabular}
\end{table*}

\subsection{Analysis and Observations}

Based on the experimental results, we can make the following observations:

\begin{enumerate}
    \item GPT-4 generally achieves higher accuracy scores than GPT-3.5 in more recent years (H29, H30, R01, R02, R03), whereas GPT-3.5 shows stronger performance in earlier years (H18, H19, H26).
    \item There is a noticeable fluctuation in the performance across years for both models, indicating that the distribution of the dataset might play a critical role in their performance on COLIEE Task 4.
    \item Interestingly, both models consistently struggle with some years such as H22 and H29, which may have a more complicated set of legal questions or a misaligned distribution of data compared to the models' training data.
\end{enumerate}

These observations offer valuable insights into the strengths and weaknesses of GPT-3.5 and GPT-4 in handling legal textual entailment tasks. Additionally, the results emphasize the importance of carefully considering training data distribution while developing specialized models for legal information extraction and entailment applications.

\section{Detailed Analysis of Experimental Results}



When comparing the average accuracy of GPT-3.5 and GPT-4 in both languages, we found a noticeable improvement in GPT-4's performance over GPT-3.5. GPT-4 achieved an average accuracy of 0.7670 in English and 0.7843 in Japanese, while GPT-3.5's average accuracy was 0.7109 for English and 0.6412 for Japanese. This indicates that the GPT-4 model offers superior performance in handling both English and Japanese datasets.


From the accuracy plots shown in Figures \ref{fig:acc_gpt35_eng_jp} and \ref{fig:acc_gpt4_eng_jp}, it is evident that both GPT-3.5 and GPT-4 models perform better with English data compared to Japanese data. However, the performance gap between the two languages is significantly reduced with GPT-4, potentially indicating its improved ability to handle Japanese language processing.


The accuracy plots in Figures \ref{fig:acc_english} and \ref{fig:acc_japanese} also reveal that the performance of both GPT-3.5 and GPT-4 models varies across years, without displaying a clear and stable upward trend. We observed fluctuations in model accuracy for certain years, such as H20, H25, and H28 for GPT-3.5 with English data, and H21 and H28 for GPT-3.5 with Japanese data. GPT-4's accuracy improved during years H20, R02, and H25 for English data, and H25, R02, and R03 for Japanese data.

This temporal variation in model performance may be partially attributed to changes in dataset difficulty and model training methods over time, as well as potential cutoffs in data availability, not only in recent years (considering GPT models' inability to predict future events) but also in the past (as GPT models may have limited understanding of past information).


One assumption we can make based on the observed trends is that the quality and availability of data in the earlier years (H18-H21) were not as standardized and publicly accessible as in recent times. This may have led to challenges in training the GPT models effectively, resulting in the relatively lower performance of GPT-3.5 during these years and somewhat similar performance between both GPT-3.5 and GPT-4 during this period.

The experimental results show an improvement in GPT-4's performance over GPT-3.5 for both English and Japanese datasets. The performance gap between the two languages has been narrowed with GPT-4, indicating its enhanced capability to process the Japanese language. Temporal variations in model performance suggest possible changes in dataset difficulty and model training methods over time, as well as potential data availability limitations in both recent and past years.

\section{Implications}

The preliminary results reveal several intriguing observations, providing a foundation for in-depth discussion on the capabilities and limitations of GPT-3.5 and GPT-4 concerning legal textual entailment tasks. 

While the findings in the ``GPT-4 Pass the Bar Exam'' paper~\cite{katz2023gpt} suggest that GPT-4 consistently outperforms GPT-3.5, our experimental results offer a more comprehensive perspective. In our evaluation across various years of the COLIEE dataset, GPT-4 does not consistently outshine GPT-3.5. 
GPT-3.5 appears to perform relatively better on older years' data compared to GPT-4. This observation raises questions related to the existence of a data cut-off not only for recent years but also for earlier periods. The balance and distribution of training data may play a crucial role in determining the performance of these models on various legal textual entailment tasks. Despite the closed nature of these models, this observation is useful for alternative open-source solutions in the future.


\section{Conclusions}

In this study, we analyzed and compared the performance of GPT-3.5 (ChatGPT) and GPT-4 in legal textual entailment tasks using the COLIEE Task 4 dataset. Our results reveal varying performance patterns depending on the context and nature of the legal questions, which raises questions about the generalizability of GPT models and their ability to learn adaptable rules for unknown cases. Subsequent research will focus on a deeper analysis of the types of questions that pose difficulties for these models, in order to better understand their limitations and devise strategies for enhancing their capabilities in handling complex legal information extraction and entailment tasks. 
Additionally, future research will also focus on challenging these models in terms of their ability to explain the reasoning behind their results. This will enable a better understanding of how the models arrive at their conclusions and facilitate improvements to their interpretability and transparency, which are essential for practical applications in the legal domain. By understanding the decision-making process of GPT models, researchers can work towards developing more accountable and trustworthy AI systems for legal textual entailment and related tasks.

\section*{Acknowledgements}
This work was supported by JSPS KAKENHI Grant Number, JP22H00543 and JST, AIP Trilateral AI Research, Grant Number JPMJCR20G4.
Francesca Toni also acknowledges support from the European
Research Council (ERC) under the European Union’s Horizon 2020 research and innovation programme (grant agreement No.101020934, ADIX), as well as support from J.P. Morgan and the
Royal Academy of Engineering, UK, under the Research Chairs and Senior Research Fellowships scheme.

\bibliographystyle{abbrvnat}
\bibliography{ref}

\end{document}